\documentclass[acmsmall]{acmart}

\usepackage{natbib}
\usepackage{subfig}
\usepackage{array}
\usepackage{hyperref}

\newcommand{\LSTM}{ LSTM-MoT }

\newcommand{\EASE}{ EASE }

\newcommand{\BERT}{ BERT }

\newcommand{\MEMORY}{ Memory-Nets }

\newcommand{\SKIPFLOW}{ SKIPFLOW }

\AtBeginDocument{%
  \providecommand\BibTeX{{%
    \normalfont B\kern-0.5em{\scshape i\kern-0.25em b}\kern-0.8em\TeX}}}

\acmConference[]{}
\acmBooktitle{}
\acmPrice{15.00}
\acmISBN{978-1-4503-XXXX-X/18/06}

\begin{document}

\title{Evaluation Toolkit For Robustness Testing Of Automatic Essay Scoring Systems}

\author[]{Anubha Kabra\textsuperscript{2*}, Mehar Bhatia\textsuperscript{1*}, Yaman Kumar Singla\textsuperscript{1,2,3}}
\authornote{Equal Contribution}
\author[]{Junyi Jessy Li\textsuperscript{4}, Rajiv Ratn Shah\textsuperscript{1}}
\affiliation{\\
  \institution{1. IIIT-Delhi, 2. Adobe, 3. SUNY-Buffalo 4. University of Texas at Austin}
  \country{1. India, 2. India, 3. USA, 4. USA}
}

\begin{abstract}

Automatic scoring engines have been used for scoring approximately fifteen million test takers in just the last three years. This number is increasing further due to COVID-19 and the associated automation of education and testing. Despite such wide usage, the AI based testing literature of these `intelligent' models is highly lacking. Most of the papers proposing new models rely only on quadratic weighted kappa (QWK) based agreement with human raters for showing model efficacy. However, this effectively ignores the highly multi-feature nature of essay scoring. %There have been a few studies noting that the scores given by AES models correlate heavily with features like length and wordiness of the essay. However, automatic essay scoring being a multifeature task 
Essay scoring depends on features like coherence, grammar, relevance, sufficiency, vocabulary, \textit{etc}., and till date, there has been no study testing Automated Essay Scoring (AES) systems holistically on all these features. With this motivation,  %Inspired by student behaviour during examinations as reported in recent news reports, 
we propose a model agnostic adversarial evaluation scheme and associated metrics for AES systems to test their natural language understanding capabilities and overall robustness. We evaluate the current state-of-the-art AES models using the proposed scheme and report the results on five recent models. These models range from feature-engineering based approaches to the latest deep learning algorithms. We find that AES models are highly overstable such that even heavy modifications (as much as 25\%) with content unrelated to the topic of the questions does not decrease the score produced by the models. On the other hand, unrelated content, on average, increases the scores, thus showing that the models' evaluation strategy and rubrics should be reconsidered. We also ask 200 human raters to score both an original and adversarial response to see if humans are able to detect differences between the two and whether they agree with the scores assigned by autoscorers.
\end{abstract}

%%
%% The code below is generated by the tool at http://dl.acm.org/ccs.cfm.
%% Please copy and paste the code instead of the example below.
%%


\maketitle

\section{Introduction}

\label{sec:Introduction}

% General Related Work

We know that writing is a social practice. Testing of written prose is a long-established practice to teach students on how to engage with readers meaningfully. It involves choosing a stance on a continuum, responding, interacting, and sharing meaning with others. Automated Essay Scoring (AES), by proposing to automate the above process, poses as an important socio-technical system in the education paradigm \citep{tang2017automated}. AES uses computer programs to automatically characterize the performance of examinees on standardized tests involving writing prose. \textit{ETS}, the largest company working in language testing domain, says that the AES systems are dependent on a balance between \textit{``current societal expectations and the cutting edge of technological advances'' }\citep{etsCompose}. The business motivation of using such systems is quite clear. They help in realizing cost-saving at scale. A human teacher is able to save hundreds of man hours per year on account of savings in testing and evaluation \cite{bryant2020artificial}. Additionally, for low-resource countries and rural areas with abysmal teacher-student ratios, this becomes a necessity \cite{ChinaOpenAES,teacherStudent}.

In the last decade, owing to the advancements in artificial intelligence, the usage of such systems has increased by several folds. These are now increasingly used in taking high-stake decisions such as college admissions, visa approvals, and job screening and pre-screening tests. In the last five years, they have further made their way to the middle and high school classrooms of states like Utah \cite{utahCompose}, and Ohio \cite{ohioAES}. While earlier, each score generated by the AI systems was verified by an expert human rater, now they are scoring a majority of essays independently without any intervention by human experts \cite{ohioAES}. At the same time, there have been a multitude of papers in premiere machine learning conferences reporting novel models and state-of-the-art on automatic essay scoring datasets \cite{ke2019automated}. The pearson-correlation based agreement scores reported by these studies have risen from 0.23 to 0.8 over time \cite{ke2019automated}. Most of these papers report Pearon-correlation or kappa based agreement scores to measure the performance of their models. However, as shown by multiple previous studies, despite achieving human level agreement scores \cite{kumar2019get} or even `surpassing' them \citep{shermis2012contrasting}, the models are easily fooled \cite{perelman2014state,perelmanBable,perelmanBableWebsite,parekh2020my}. This reduces the trustworthiness of AI-based automated scoring systems in the eyes of both language-testing researchers \cite{perelman2014state,parekh2020my,west2018trustworthy,reinertsen2018can} and general public \cite{ohioAES,greene2018automated,viceFlawed,smith2018more}.

Due to its wide applicability, several research studies in the linguistic community have tried to characterize the performance of essay scoring models and attribute it to features like number of words \cite{perelman2014state}, style \cite{reinertsen2018can}, vocabulary \cite{perelmanBableWebsite}, coherence \cite{ding2020don}, \textit{etc.} However, the results from these research studies are often conflicting in nature. While one indicates that essay scoring models have substantial correlation with number of words \cite{perelman2014state}, the other attributes it to style \cite{reinertsen2018can}. Moreover, there is no standard way of testing automatic essay scoring systems apart from measuring agreement scores on a subset of the dataset (typically chosen to be 10\% of the dataset size) \cite{west2018trustworthy}. This leads to non-thorough testing and hence model development. It is noteworthy that in the last five years, very few publications have performed any evaluations beyond agreement scores. Most of those who have reported any other feature do it mostly on coherence evaluation \cite{xu2019cross,jeon-strube-2020-centering,tay2018skipflow}. This is inadequate evaluation technique since essay scoring is a highly feature rich task which depends on a variety of features like vocabulary, factuality, coherence, grammar, relevance, sufficiency, argument quality, persuasion, \textit{etc.} \cite{yan2020handbook}.

Despite the importance and the magnitude of the problem, there have been a few efforts from language testing community to develop a unified testing framework. Ding \textit{et al.} \citep{ding2020don} collaborated with \textit{ETS} \footnote{https://www.ets.org/} to show that AES models are adversarially perturable. However, the inputs are limited to just random incoherent response generation. This does not mimic a test-taker's capability to fool an AES system nor does it test all the features important for scoring. There have also been some manual studies where experts and non-experts were invited to test out some models \citep{powers2001stumping}. These studies, despite the good motivation and human grounding efforts, cannot be scaled or even made consistent across all the models. To the best of our knowledge, there has been no work which systematically analyzes AES models on all the different aspects important for scoring or propose an evaluation suite. Such a validity suite is important from the following perspectives: 1)~it provides a uniform benchmark to compare different models beyond metrics such as accuracy or QWK. These metrics neither provide any insights into the construct validity of AES models nor do they indicate the robustness of a model, 2)~it builds trust in the automatic scoring system, and 3)~it promotes understanding of the black-box AES models.

\subsection{Why Agreement Scores Based On Quadratic Weighted Kappa (QWK) And Pearson Correlation Are Inadequate Performance Metrics?}
\label{sec:Why agreement scores based on quadratic weighted kappa (QWK) and Pearson Correlation are inadequate performance metrics?}
The common performance metric that has been widely used in the field is Quadratic Weighted Kappa (QWK). It measures the agreement between the scoring model and the human expert. Given observed scores matrix $O$ (confusion scores), weights $w$ (containing penalty of each possible predicted score with each possible actual score) and expected score matrix $E$, number of possible scores $N$, QWK is defined as:
%%%%%%%%%%%
\begin{equation}
\small
 k = 1 - \Sigma_{ij} w_{ij} O_{ij} / \Sigma_{ij} w_{ij} E_{ij}
\end{equation}

$O_{ij}$ measures number of students who received a score $i$ by the human grader and $j$ by the model. Weight matrix is defined as ($w_{ij} = (i-j)^2/(N-1)^2$) and assigns penalty to each pair of predicted, actual scores.
QWK denotes machine-human agreement. It is then compared with human-human agreement score to compare different models.
%%%%%%%%%

The other metric commonly used in the literature is Pearson Correlation (PC). Given $N$ as the number of pairs of scores, $\Sigma x y$ as the product of paired scores, $\Sigma x$ and $\Sigma y$ being the sum of x and y scores respectively and $ \Sigma x^{2}$ ,$\Sigma y^{2}$ referring to the sum of the squares of x and y scores.  It is defined as:
%%%%%%%%
\begin{equation}
r=\frac{N \Sigma x y-(\Sigma x)(\Sigma y)}{\sqrt{\left[N \sum x^{2}-(\Sigma x)^{2}\right]\left[N \Sigma y^{2}-(\Sigma y)^{2}\right]}}
\end{equation}
%%%%%%%%

We argue that for deep learning based systems, tracking merely QWK (or PC) as evaluation metrics is suboptimal for several reasons: 1)~while subsequent research papers show an iterative improvement in QWK but most of them fail in evaluating how their works generalize across all the different dimensions of scoring including coherence, cohesion, vocabulary, and even surface metrics like average length of sentences, word difficulty, \textit{etc}. 2)~QWK as a metric captures only the overall and broad agreement with humans scores, however, scoring as a science includes knowledge from many domains of NLP like: \textit{fact-checking, discourse and coherence, coreference resolution, grammar, content coverage, \textit{etc}} \citep{yan2020handbook}. %A neural network normally tries to learn all of them at one go, which as the results demonstrate is probably not able to learn. 
QWK, instead of making the scoring comprehensive, is abstracting out all the details associated with scoring as a task. 3)~it does not indicate the direction of a machine learning model: oversensitivity or overstability.
We quantitatively illustrate the gravity of all these aspects by performing statistical and manual evaluations, mentioned in Section~\ref{sec:adversarial-evaluation}.

We demonstrate in the later parts of our paper that heavily modifying responses (as much as 25\%), does not break the scoring systems and the models still maintain their high confidence and scores while evaluating the adversarial responses. Our results show that no published model is robust to these examples. They largely maintain the scores of the unmodified original response even after all the adversarial modifications. This indicates that the models are largely overstable and unable to distinguish ill-formed examples from the well-formed ones. While on an average, humans reduce their score by approx 3-4 points (on a normalized 1-10 scale), the models are highly overstable and either increase the score by 1 point for some tests or reduce them for others by only 0-2 points (\S~\ref{sec:human-annotation-survey}).
We propose that instead of tracking \textit{just} QWK for evaluating a model, the field should track a combination of QWK and adversarial evaluation of the models for performance.

\subsection{Basis Of The Evaluation Suite}
\label{sec:Basis Of The Evaluation Suite}
Cognitive studies have characterized AES models as information-integration models trying to learn category-learning tasks \citep{yan2020handbook}. The descriptor of such a category can be, \textit{``Score the essay at level 3 if it consists of a clear aim reasoned by structured claims and supported by appropriate evidence with rebuttals of all the major counter arguments."} \citep{yan2020handbook}. Following this, many research studies have established features which must be present in AES models \citep{yan2020handbook,burstein2004automated,sukkarieh2009c,kumar2019get}. A few examples of such features are: factuality, grammar-correctness, organization, coherence, lexical sophistication, \textit{etc}. In this work, we propose a black-box adversarial evaluation of AES systems based on these features. We show the evaluation of five recent models on the popular dataset, Automated Student Assessment Prize (ASAP) dataset for Essay-Scoring \citep{ASAP-AES}. 

Our evaluation scheme consists of evaluating AES systems on essays derived from the original responses but modified heavily to change its original meaning. These tests are mostly designed to check for the overstability of the different models. An overview of the adversarial scheme is given in Table~\ref{table:overall-test-case-list}. We perform the following operations for generating test responses: \textit{Addition} (Adding lines to the original text), \textit{Deletion}(Deleting lines from the original text), \textit{Modification}(Modifying parts of the original text) and \textit{Generation}(Generating a completely new text). 

These cover all the fundamental methods that can be used to change a given piece of sequence to another \cite{martin1983managing}. Under these four operations, we include many other operation subtypes such as adding related and unrelated content, modifying the grammar of the response, taking only first part of the response, \textit{etc}. These operations and sub-operations quantify a model's performance on each feature important for scoring.

Therefore, the main contributions of our work are summarized as follows:
\begin{itemize}
 \item We propose a model agnostic evaluation suite to alter examples given in a dataset to test out a given AES model. This evaluation suite can be used to test various systems including automatic scoring \cite{easeGithub,tay2018skipflow}, attribute scoring \cite{mathias-bhattacharyya-2018-asap}, coherence evaluation \cite{jeon-strube-2020-centering}, argument mining \cite{nguyen-2018-comparing}, topic detection \cite{yang2018topic}, and measuring argument persuasiveness \cite{ke2018learning}. Essay scoring datasets like the ASAP-AES dataset were used in all these settings and hence our evaluation suite can also be used in all these settings.
 
 \item We evaluate five recent state-of-the-art AES models on all the eight prompts belonging to the widely-cited \citet{ASAP-AES} dataset and report their test performance on various metrics for a thorough understanding of their weaknesses.
 
 \item We propose a comprehensive 3-way automatic evaluation for aiding model-makers involving parameters of length, position and type of adversarial tests. We also validate the adversarial examples with a human study to show that scores awarded by AES models are indeed disconnected with rubrics.
 
 \item Finally, we open-source the code, test samples and model weights for easy reproducibility, and future benchmarking.
\end{itemize}

We would also like to say that we present our argument not as a criticism of anyone, but as an effort to refocus the research directions of the field. Since the automated systems that we develop as a community have such high stakes like deciding jobs and admissions of the takers, the research should reflect the same rigor. We sincerely hope to inspire higher quality reportage of the results in automated scoring community that does not track just the performance but also the validity of their models.

\section{Task and Setup}
\label{section:taskAndSetup}
% In this section, we discuss

In this section, we define the problem statement and the dataset used for experimentation. We provide details about the state-of-the-art AES models we experimented with and the adversarial evaluation metrics. We also elaborate on all the adversarial test cases used for testing these models.
\subsection{Task and Dataset}
\label{subsec:Task and Dataset}

Similar to various research studies \citep{taghipour2016neural,easeGithub,tay2018skipflow,zhao2017memory}, we have used the widely cited ASAP-AES \citep{ASAP-AES} dataset to evaluate Automatic Essay Scoring systems.
%ASAP-SAS has been used for shorter ones \citep{kumar2019get,riordan2017investigating,ramachandran2015identifying,liu2019automated}. 
%It is one of the most extensive publicly available datasets. 
The relevant statistics for this dataset are listed in Table~\ref{table:AES-dataset-stats}. 
%and~\ref{table:SAS-dataset-stats}. 
The questions covered by the dataset are from many different areas such as Sciences and English literature. The responses were written by high school students and were subsequently double-scored. The evaluation framework built for assessing AES systems is broadly based on the linguistic features considered essential for scoring like grammar, coherence, \textit{etc} \citep{bejar2017automated,yan2020handbook}.
% \vspace*{-\baselineskip}
\begin{table}[!ht]
\footnotesize
\centering
%\resizebox{\textwidth}{!}{
\begin{tabular}{lllllllll}
\hline
\textbf{Prompt Number} & \textbf{1} & \textbf{2} & \textbf{3} &\textbf{ 4} & \textbf{5} &\textbf{ 6 } &\textbf{ 7 } & \textbf{8} \\\hline
\#Responses & 1783 & 1800 & 1726 & 1772 & 1805 & 1800 & 1569 & 723 \\
Score Range & 2-12 & 1-6 & 0-3 & 0-3 & 0-4 & 0-4 & 0-30 & 0-60 \\
\#Avg words per response & 420 & 430 & 127 & 109 & 147 & 180 & 205 & 710 \\
\#Avg sentences per response & 23 &20 &6 &4.5 &7 &8 &12 &35 \\ 
% \#Type &Argumentative & Argumentative &Reading Comprehension &Reading Comprehension &Reading Comprehension &Reading Comprehension & Narrative &Narrative
Type & Ar & Ar & RC & RC & RC & RC & Na & Na\\ \hline
% Topic &Computers &Censorship in the Libraries &Rough Road Ahead &Winter Hibiscus &Home: The Blueprints of Our Lives &The Mooring Mast &Patience &Laughter\\\hline
\end{tabular}
%}
\caption{\small \label{table:AES-dataset-stats} Overview of the ASAP AES Dataset used for evaluation of AES systems. \small (RC = Reading Comprehension, Ar = Argumentative, Na = Narrative).}
\end{table}

% \vspace*{-\baselineskip}
\vspace{-6mm}
\subsection{Models}
\label{subsection:Models}
We evaluate the recent state-of-the-art deep learning \citet{taghipour2016neural,tay2018skipflow,zhao2017memory,liu2019automated} and feature-based models \citet{easeGithub} and show the adversarial-evaluation results. Brief descriptions of each of them are given as follows: 
\begin{itemize}
    \item \textbf{\EASE } (\citet{easeGithub}): It is an open-source feature-based model maintained by \textit{EdX}. This model includes features such as tags, prompt-word overlap, n-gram based features, \textit{etc}. Originally, it ranked third among the 154 participating teams in the ASAP-AES competition.
    \item \textbf{\LSTM} (\citet{taghipour2016neural}): They use CNN-LSTM based neural networks with a few mean-over-time layers to score essays. The paper reports 5.6\% improvement of QWK on top of the \textit{\EASE} feature-based model.
    \item \textbf{\SKIPFLOW} (\citet{tay2018skipflow}): \SKIPFLOW provides a deep learning architecture that captures coherence, flow and semantic relatedness over the length of the essay, which the authors call \textit{neural coherence features}. SkipFlow accesses intermediate states to model longer sequences of essays. Doing this, they show an increase of 6\% over EASE feature engineering model and 10\% over a vanilla LSTM model. 
    \item \textbf{\MEMORY} (\citet{zhao2017memory}): The authors use memory-networks for automatic scoring where they select some responses for each grade. These responses are stored in the memory and then used for scoring ungraded responses. The memory component helps to characterize the various score levels similar to what a rubric does. They compare their results with the EASE based model and show better performance on 7 out of 8 prompts.
    \item \textbf{\BERT} (\citet{liu2019automated}): This work makes use of adversarial examples to improve AES. They consider two types of adversarial evaluation: well-written permuted paragraphs and prompt-irrelevant essays. For these, they develop a two-stage learning framework where they calculate semantic, coherence and prompt-relevance scores and concatenate them with engineered features. The paper uses BERT \citep{devlin2018bert} to extract sentence embeddings. 
\end{itemize}
%%%%%%%%%%%%%%%%%%%%%%%%%%%%%%%%%%%%%%%%%%%%%%%%%%
%%%%%%%%%%%%%%%%%%%%%%%%%%%%%%%%%%%%%%%%%%%%%%%%%%

% \subsection{Standard Evaluation}
% \label{sec:Standard Evaluation}

% Quadratic Weighted Kappa is the standard metric for evaluating model performance on the essay scoring task \citep{attali2004automated}. Both the competition which released the ASAP-AES dataset and the subsequent papers using that employ it as the evaluation metric. Given observed scores matrix $O$ (confusion scores), weights $w$ (containing penalty of each possible predicted score with each possible actual score) and expected score matrix $E$, number of possible scores $N$, QWK is calculated as 
% \begin{equation}
%  k = 1 - \Sigma_{ij} w_{ij} O_{ij} / \Sigma_{ij} w_{ij} E_{ij}
% \end{equation}

% $O_{ij}$ measures number of students who received a score $i$ by the human grader and $j$ by the model. Weight matrix is defined as ($w_{ij} = (i-j)^2/(N-1)^2$) and assigns penalty to each pair of predicted, actual scores.
% QWK denotes machine-human agreement. It is then compared with human-human agreement score to compare different models.

%%%%%%%%%%%%%%%%%%%%%%%%%%%%%%%%%%%%%%%%%%%%
%%%%%%%%%%%%%%%%%%%%%%%%%%%%%%%%%%%%%%%%%%%%

\subsection{Evaluation Framework}
\label{sec:adversarial-evaluation}

\subsubsection{General Framework}
\label{sec:general-framework}
\begin{figure}[!ht]
 \centering
 \includegraphics[width=7cm]{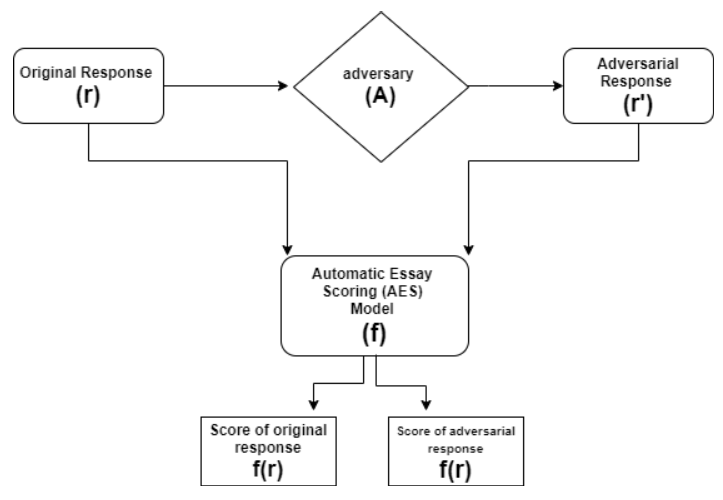}
 \caption{General Framework for Test Evaluation Model given a prompt $p$}
 \label{general framework}
\end{figure}

From Figure \ref{general framework}, we can see that given a prompt $p$, response $r$, bounded size criterion $c_1$, position criterion $c_2$ and optionally a model $f$, an adversarial testing model $A$ converts response $r$ to response $r'$ based on a specific set of rules and the criteria $c_1$ and $c_2$. %The criterion $c_1$ defines whether the size of the response should change after adversarial perturbation or not. 
% \begin{figure}[!ht]
%  \centering
%  \includegraphics[width=9cm]{framework4.PNG}
%  \caption{General Framework for Adversarial Evaluation given a prompt $p$}
%  \label{general framework}
% \end{figure}

The criterion $c_1$ defines the percentage upto which the original response has to be changed by the adversarial perturbation such that 
$|Len(r')~-~Len(r)|/Len(r)~=~c_1$. We try out different values of $c_1$ (\{10\%, 15\%, 20\%, 25\%\}). The criterion $c_2$ defines the position of inducing adversarial perturbation. We consider three positions ($\{\textsc{Start, Mid, End}\}$) by dividing the response $r$ into three equal-sized portions. The results are presented in the Section~\ref{sec:Effect of Choice of Different Hyperparameters}.

\begin{table}[ht]
%\small
\centering
\resizebox{\columnwidth}{!}{
\begin{tabular}{lll}
% \begin{tabular}{|m{0.1\columnwidth}|m{0.35\columnwidth}|m{0.45\columnwidth}|}
\hline
Symbol & Name & Defn.\\ \hline

$N_{pos}$& Percentage of positively impacted samples & \# $r/N$ s.t., $f(r) < f(r')$) \\
$N_{neg}$& Percentage of negatively impacted samples & \# $r/N$ s.t., $f(r) > f(r')$) \\

$\sigma$ & Standard deviation of the difference & $\sqrt{\Sigma (f(r) - f(r') - \mu)^2/N}$\\
$\mu_{pos}$ & Mean difference in scores for positively impacted samples & $\Sigma (f(r')-f(r))/N$ s.t. $f(r) < f(r')$\\
$\mu_{neg}$ & Mean difference in scores for negatively impacted samples & $ \Sigma (f(r)-f(r'))/N$ s.t. $f(r) > f(r')$ \\ \hline
\end{tabular}}
%}
% \vspace{-4mm}
% \vspace*{-\baselineskip}
\caption{Adversarial Evaluation Metrics. $(r, r')$ denote the original (human-written) and adversarial samples and $f(r), f(r')$ denote the score of an automatic scoring model on original and adversarial response. $N$ is the size of universal set of test responses.}
\label{table:adversarial-eval-metrics}
\end{table}
% \vspace{-5pt}

% $\mu$ & Mean difference & $\Sigma (f(r)-f(r'))/N)$ \\ $\mu_{abs}$ &Absolute mean difference& $|\Sigma (f(r)-f(r'))|/N$\\

For benchmarking a model $f$, we use the scores $f(r)$ and $f(r')$ to calculate the statistics listed in Table~\ref{table:adversarial-eval-metrics}. Since the score ranges and the number of samples vary across all the prompts, we report the corresponding values in percentages (percentage of total samples and percentage of range of score). From our human evaluation survey  (Section~\ref{sec:human-annotation-survey}) and corresponding Table \ref{table:HumanAnnotations}, we see a significant difference in human scores and scores generated by various AES systems. 
We ask our human annotators to score our adversarial response \textit{r'}, given the score for the original 
response \textit{r}. We also ask the annotators to give supporting reasons for their responses. From our survey, for each adversary, $A$, we summarize the following,  
\begin{enumerate}
 \item According to all human annotators, the score of an adversarial response ($r'$) was always less than the score of the original response ($r$). In other words, from humans' point of view, no adversary increased the quality of the response. 
 \item Second, all human annotators were able to detect and differentiate $r$ from $r'$. We conducted t-test between scores given by AES engines and human annotators on the adversarially perturbed responses to confirm this notion. 94\% of all the t-tests rejected the null hypothesis (\textit{p<0.05}), hence highlighting the statistical significance.
\end{enumerate}

Notably, these findings are different from what is ``commonly'' given in the adversarial literature where the adversarial response is formed such that a human is not able to detect any difference between the original and modified responses, but a model (due to its adversarial weakness) is able to detect differences and thus changes its output \citep{zhang2020adversarial}. For example, in computer vision, a few pixels are modified to make the model mispredict a bus as an ostrich \citep{szegedy2013intriguing}, and in NLP, paraphrasing by changing a few words is done to churn out racial and hateful slurs from a generative deep learning model \citep{wallace2019universal}. Here, our survey observations show that humans can \textit{detect} the difference between the original and final response. We call the inability (or under-performance) of models on differentiating between adversarial and natural samples as their \textit{overstability}.

\begin{table}[!ht]
\centering
%\begin{adjustbox}{\textwidth}
\footnotesize
\scalebox{0.9}{
\begin{tabular}{|m{0.01\columnwidth}|m{0.1\columnwidth}|m{0.17\columnwidth}|m{0.7\columnwidth}|}
% \begin{tabular}{llll}

\hline \textbf{\#} & \textbf{Category} & \textbf{Test Name} & \textbf{Description} \\ \hline
1 & \textsc{Add} & \textsc{AddWikiRelated} & Addition of Wikipedia lines related to the essay question in a response.\\

 & & \textsc{AddWikiUnrelated} & Addition of Wikipedia lines unrelated to the essay question in a response. \\
 & & \textsc{RepeatSent} &Repetition of some lines of the response within a response. \\
 & & \textsc{AddSong} & Addition of song lyrics into the response.\\
 & & \textsc{AddSpeech} & Addition of excerpts of speeches of popular leaders into a response.\\
 & & \textsc{AddRC} & Addition of lines from Reading Comprehension based questions into a response. \\
 & & \textsc{AddTruth} & Addition of True lines into a response.\\
 & & \textsc{AddLies} & Addition of Universally false lines into a response.\\ \hline

 2 & \textsc{Delete} & \textsc{DelStart} & Deletion of lines from the beginning of a response.\\
 & & \textsc{DelEnd} & Deletion of lines from the end of a response. \\

& & \textsc{DelRand} & Deletion of random lines from a response. \\ \hline

3 & \textsc{Modify} & \textsc{ModGrammar} & Modifying the sentences in a response to have incorrect grammar.\\
 %& & \textsc{ModFluency} & Inducing dis-fluency in the sentences of a response. \\

 & & \textsc{ModLexicon} & Paraphrasing words in the sentences with their respective synonyms in a response. \\ 
 & & \textsc{ShuffleSent} & Randomly shuffling the sentences in a response.\\\hline

4 & \textsc{Generate} & \textsc{BabelGen} & Using the essay generated by \textit{Babel} as a response.\\ \hline
\end{tabular}}
%\end{adjustbox}
\caption{
\small
\label{table:overall-test-case-list} Overview of the testing scheme for Automatic Essay Scoring (AES) models.}
\end{table}

Next, we discuss the various strategies of adversarial perturbations. An overview of all the perturbations is given in Table~\ref{table:overall-test-case-list}. We categorize all the adversarial tests by the \textit{major}-operation they do on a sample. Therefore, we divide the tests into four categories: \textsc{Add} (those operations which change a sample majorly by adding to it), \textsc{Delete} (those operations which change a sample majorly by deleting from it), \textsc{Modify} (those operations which change a sample majorly by modifying its structure) and \textsc{Generate} (those operations which tests the robustness of a model by giving it completely machine-generated non-meaningful samples).

\subsubsection{\textbf{\textsc{Add} Adversaries}}
\label{sec:Add Adversaries}

\textsc{Add} adversaries change the original response by adding new content to it. Adding unrelated or repetitive content negatively impacts the content-specific and topic development features of an essay, which are considered necessary for essay evaluation \citep{yan2020handbook}. To test the content knowledge of scoring models, we designed various types of \textsc{Add} tests that are explained hereafter.All the testcases follow the position and amount of addition given by the parameters, $c_1$ and $c_2$, respectively, as explained in Section~\ref{sec:general-framework}. A few examples are shown in Figure \ref{add_examples}.

%\subsubsection{\textsc{AddWikiRelated}}

\begin{itemize}
 \item \textsc{\textbf{AddWikiRelated}}: With this testcase, we add prompt-related information to each sample response. We used a key-phrase extraction technique~\footnote{https://github.com/boudinfl/pke} over each prompt/question in the dataset for choosing prompt-related articles from Wikipedia\footnote{https://pypi.org/project/wikipedia/}. After selecting articles, we randomly selected sentences from each extracted article and appended them to the responses.  

\item {\textsc{\textbf{AddWikiUnrelated}}}: 
%It is important to consider topic-development features when scoring an essay. These features represents the main ideas of a test-taker and measuring the meaning of the response. The important measures to note are essay content and discourse \citep{yan2020handbook}. 
We form this testcase to disturb the topic relevance of the responses. This test tries to mimic students' behavior when they make their response lengthy by adding irrelevant information. For this, we add prompt-irrelevant information to each sample response by selecting Wikipedia articles that do not match the response's prompt. The score by an AES model should be negatively affected with this kind of perturbation. The first example in Figure~\ref{add_examples} depicts this testcase.

\begin{figure}[!ht]
 \centering
 \includegraphics[width=\textwidth]{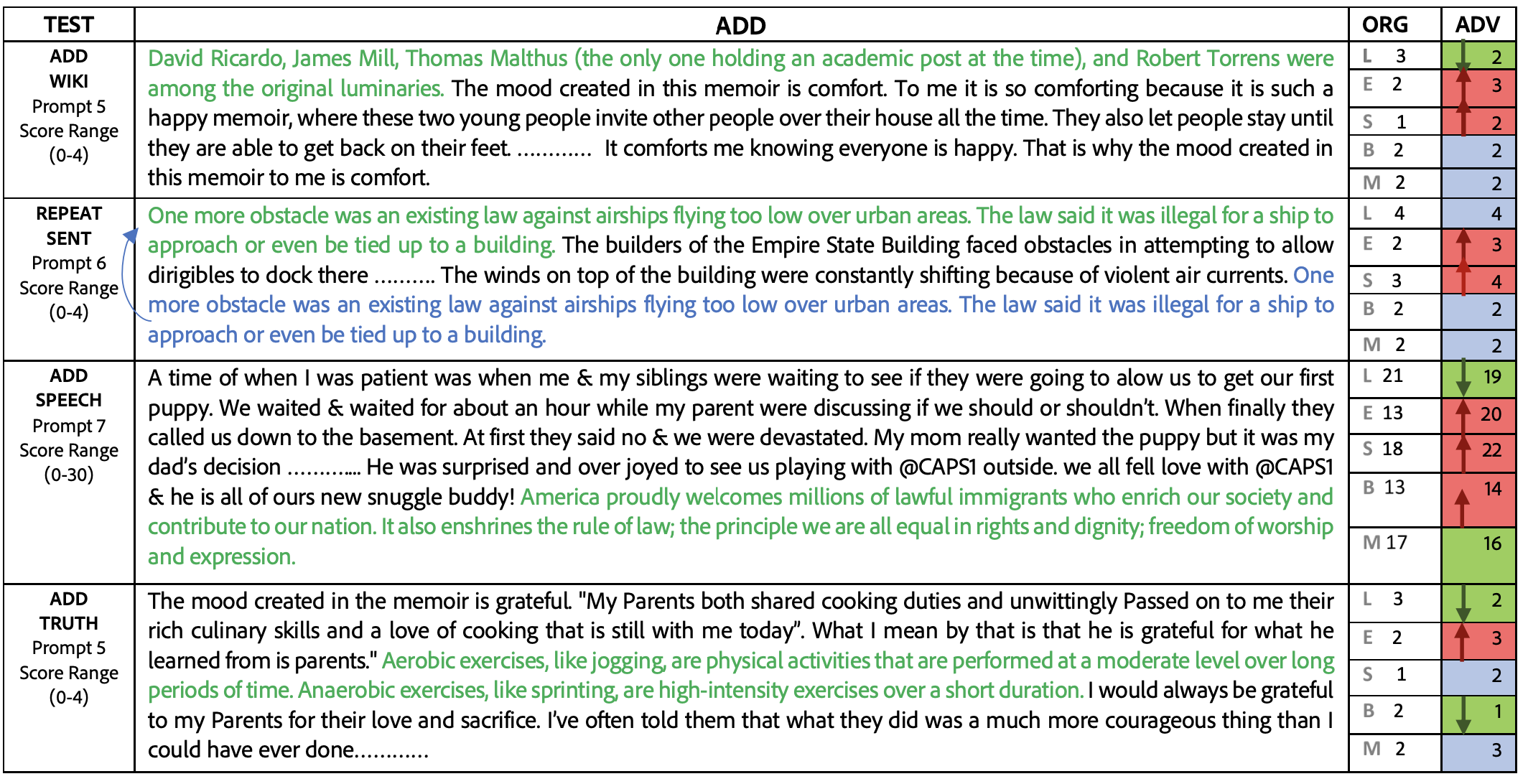}
\caption{Examples of the \textsc{ADD} testcases. Here, ORG refers to the score given to the original sample and ADV refers to the score given to the adversarial sample by each model. The list of models according to which scores are listed is LSTM (L), EASE (E), SKIPFLOW(S), BERT (B) and MEMORY NETS (M). \textcolor{red}{Red}, \textcolor{green}{Green}, \textcolor{blue}{Blue} shows that adversarial responses were scored \textcolor{red}{higher}, \textcolor{green}{lower} and \textcolor{blue}{equally} than original response, respectively.}
 \label{add_examples}
\end{figure}

\item \textsc{\textbf{RepeatSent}}: Students intentionally tend to repeat sentences or specific keywords in their responses in order to make it longer yet not out of context and to fashion cohesive paragraphs \citep{higgins2014managing,lochbaum2013detection,yoon2018atypical}. This highlights the test taker's limited knowledge about the subject and also clutters the writing. To design responses for this test, we divided each response into three equally sized chunks and randomly selected sentences from each of them to form a repetition block, added back to the response. An AES model should negatively score such responses. The second example in Figure~\ref{add_examples} depicts this testcase.

\item \textsc{\textbf{AddSong}}: Poetic license gives freedom to ignore or modify normal English rules. However, creative content like songs have a very different language structure than written prose in tests. Therefore, this can be used for negative testing of a system. Additionally, it has been observed that students in an attempt to fool the system use this strategy in their exams \citep{midDay}. With this motivation, we form this test by perturbing samples to include songs. We used \citet{songsKaggle4,songsKaggle3,songsKaggle2,songsKaggle5} and \citet{songsKaggle1} to extract 58,000 English songs lyrics over a long time period and range of genres like Rock, Jazz, Classical, \textit{etc}. An AES system should negatively score such responses with addition of song lyrics since they do not relate to the prompt and are a misfit to the context of the answer.

\item \textsc{\textbf{AddSpeech}}: Formal style of writing or speech is conventionally characterized by long and complex sentences, a scholarly vocabulary, correct grammatical rules and a consistently serious tone \citep{obrecht1999minimum}. In the speeches of leaders, popular terms might be used to refer to certain contextual social phenomenon. It may also include references to literary works or allusions to classical and historical figures. Generally, this style of writing is seen as sophisticated and hence better. However, when sentences of such a type are added without context or relevance, they serve the purpose of confusing the readers without giving any new meaning. We collected eight public speeches of popular leaders such as Barack Obama, Hillary Clinton, Queen Elizabeth II, \textit{etc.} These speeches were sourced from public archives and government websites.

\item \textsc{\textbf{AddRC}}: It is commonly observed that students tend to repeat parts of a question in their answer to make their answers lengthier and related to the question asked \citep{higgins2014managing,lochbaum2013detection,yoon2018atypical}. Therefore, to test over-reliance of AES models on the keywords present in a question asked or reading comprehension given, we randomly pick up sentences from the corresponding reading comprehension passages and add them to the responses.

\item \textsc{\textbf{AddTruth}}: Facts and quotations provide conclusive evidence and a voice of authority for the arguments addressed in an essay \citep{studyTipsCompose} which makes it common for test-takers to use. The motive behind this testcase is to measure relevance of responses \citep{yan2020handbook} and a check for factuality knowledge in current AES systems. This attack focuses on inculcating factual, yet unrelated text,  often done by students to increase the word count of the responses. For this testcase, we acquired a list of well-known facts from \citep{trueFacts} and injected it into the original text.

\item \textsc{\textbf{AddLies}}: Test takers may use false facts or quotations to embellish their essays and provide strong argumentative evidence to their reasoning written in their response. This underscores the importance of fact-checking while scoring these essays. This forms the motive behind this testcase and check whether these systems are able to highlight this disinformation. We collected various false statements\footnote{We used the website (https://thespinoff.co.nz/science/28-10-2017/101-fake-facts-that-youre-doomed-to-remember-as-true/)}, manually verified them to be false statements and did not include those which we felt were subjective in nature. We also note that \textsc{AddLies} being false statements should preferably impact the scoring more negatively than \textsc{AddTruth}.

\end{itemize}

% \subsubsection{\textsc{AddSong}}
% Songs and poems, because of their poetic license, are often used to substantiate certain points in an essay and show knowledge of the material.

%\subsubsection{\textsc{AddSpeech}}

%The collected speeches with their sources are given in the supplementary.\\
%\subsubsection{\textsc{AddRC}}

%\subsubsection{\textsc{DelStart}}

\subsubsection{\textbf{\textsc{Delete} Adversaries}}
\label{sec:Del Adversaries}

\textsc{Delete} adversaries change the original response by deleting content from it. These tests generally break the flow of an argument, delete crucial details from an essay and decrease wordiness. This can seriously detract from the coherency and quality of writing and frustrate readers. The various types of \textsc{Delete} tests are explained hereafter. Some examples of these tests are shown in Figure \ref{del_examples}.

\begin{itemize}
 \item \textsc{\textbf{DelStart}}: Beginnings generally serve the purpose of introducing the flow of an essay. They state the main point of the overall argument and give context to what will come in the next paragraphs. It helps in outlining a response. Hence, it is crucial to maintain the discourse of an essay and its central features like organization and development \citep{yan2020handbook}. Although organization may not be severely impacted on deleting introductory lines, the essay's development will crumple. In this testcase, we remove the introductory lines from each response which renders the development senseless, hence negatively impacting the scores. The first example in Figure~\ref{del_examples} depicts this testcase.

\begin{figure}[!h]
 \centering
 \includegraphics[width=\textwidth]{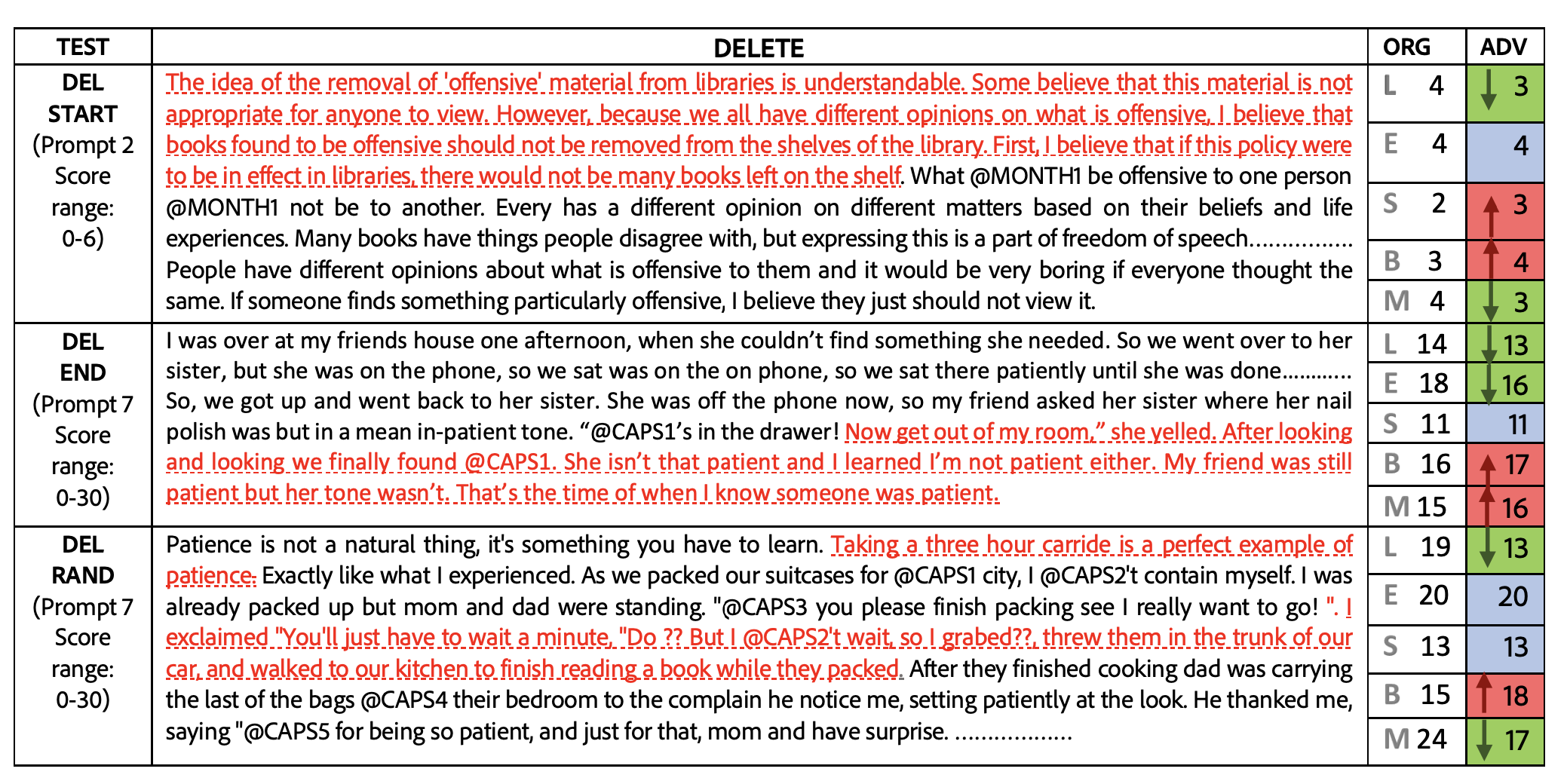}
 \caption{Examples of all the \textbf{\textsc{DEL}} testcases. Here, ORG refers to the score given to the original sample and ADV refers to the score given to the adversarial sample by each model. The red highlighted text was deleted from the original responses. The list of models according to which scores are listed is LSTM (L), EASE (E), SKIPFLOW(S), BERT (B) and MEMORY NETS (M). \textcolor{red}{Red}, \textcolor{green}{Green}, \textcolor{blue}{Blue} shows that adversarial responses were scored \textcolor{red}{higher}, \textcolor{green}{lower} and \textcolor{blue}{equally} than original response, respectively.}
 \label{del_examples}
\end{figure} 
% {\RaggedRight
% \begin{table}[!ht]
% \scalebox{0.8}{
% \begin{tabular}{|m{0.6\columnwidth}|m{0.6\columnwidth}|}
% \hline
% \textbf{Original} & \textbf{Perturbation using \textsc{DelStart}}\\
% \hline
% \textit{Example 1:}
% \textcolor{purple}{Dear newspaper, I think that people should have computers. I think that because kids play games to help them learn like math game or read and grown ups can go on websites and watch videos and do their office work.} They could also pay their bills on the computer and do shopping if they do not want to go to the store. There are many advantages of having a computer. & They could also pay their bills on the computer and do shopping if they do not want to go to the store. There are many advantages of having a computer.\\ \hline
%  \textit{Example 2:}
%  \textcolor{purple}{The setting affected the cyclist a lot. One way was the heat. It got so hot that he was becoming dehydrated. He kept drinking his water, until he didn't have any left.} Another reason is because the road turned into hills which made him use more energy. Finally, it didn't help that he hadn't seen any sign of living creatures except for a snake. & Another reason is because the road turned into hills which made him use more energy. Finally, it didn't help that he hadn't seen any sign of living creatures except for a snake.\\

% \hline
% \end{tabular}}
% \caption{Examples of perturbation to response using \textsc{DelStart}. Text marked in \textcolor{purple}{purple} highlight the deleted portion.}
% \label{tab:DelStart}
% \end{table}
% }

 \item \textsc{\textbf{DelEnd}}: Similar to the above test, we deleted the last conclusive sentences from an essay. The conclusion of any response is also an integral part of an essay. It allows you to have the final say on the arguments you have raised, synthesize your thoughts, demonstrate the importance of your ideas, and propel your reader to a new view of the subject. The conclusion is the point where the final argument is stated based on the evidence provided in the body of the essay. Deleting the conclusion, therefore, must decrease the score of the overall essay.

 \item \textsc{\textbf{DelRand}}: Organization of an essay is critical for the readers to understand the flow and context of the essay and maintain the overall cohesion. It describes how the essay holds together. The transition between one point to another should be clear and not abrupt. In summary, to disrupt the organization of an essay, we removed sentences randomly from the response. AES systems should lower the scores for these essays.

\end{itemize}

\subsubsection{\textbf{\textsc{Modify} Adversaries}}
\label{sec:Modify Adversaries}
\textsc{Modify} adversaries majorly retain the originality of a response while changing its syntax heavily. In this, we majorly change the grammar, fluency, organization and lexical sophistication of a sample. The various types of \textsc{Modify} tests are explained hereafter. Some examples of these tests are shown in Figure \ref{modify_examples}.

%\subsubsection{\textsc{ModGrammar}}
\label{sec:ModGrammar}

%\vspace{-1pt}
% \vspace*{-\baselineskip}
\begin{itemize}
 \item \textsc{\textbf{ModGrammar}}: Several studies underline the importance of grammar in scoring \citep{attali2004automated,burstein2004automated}. \textit{TOEFL iBT} mentions grammar usage in the category `language use' for their TOEFL test \citep{cushing2010validation}. We formed two test cases to simulate common grammatical errors committed by students. The first one focused on evaluating the basic grammar knowledge of AES models and the second one assessed the effect of colloquial and informal language commonly found in essays as is demonstrated in the Table~\ref{table:modGrammar}. For changing the subject-verb-object (SVO) order, we parse the responses and using spacy\footnote{https://spacy.io/} library to extract grammatical dependencies. An abbreviation dictionary~\footnote{https://abbreviations.yourdictionary.com/articles/list-of-commonly-used-abbreviations.html} is used for randomly replacing words with their corresponding informal colloquial forms. The first example in Figure~\ref{modify_examples} depicts this testcase.
\begin{table}[ht]
\small 
\centering
\begin{tabular}{l|l}
\hline
\textbf{Original} & \textbf{Anita is going to the park for a walk.} \\ \hline
Subject-Verb-Object Order Errors & Anita to the park is going for a walk. \\ \hline
Step 1: Article Errors & Anita is going to an park for the walk. \\
Step 2: Subject Verb Agreement Errors & Anita go to an park for the walk. \\
Step 3: Conventional Errors & anita go 2 an park 4 the walk \\ \hline
\end{tabular}
% \vspace{-4mm}

\caption{Examples of the type \textsc{ModGrammar}}
\label{table:modGrammar}
\end{table}
\vspace*{-\baselineskip}
\item \textsc{\textbf{ModLexicon}}: Diversity and sophistication of vocabulary is an essential feature for scoring essays \citep{chen2018end,kumar2019get}. It is commonly observed that test-takers using sophisticated vocabulary often are scored higher than their counterparts using simpler, more straightforward vocabulary \citep{perelmanBable}. However, the change or inclusion of even a single word in a sentence changes its meaning. Therefore, in this test case, we evaluate AES systems' vocabulary-dependence by improper replacement of a random word (excluding stopwords) in each sentence, to a synonym using Wordnet synsets \citep{miller1995wordnet}. Later, in Section~\ref{sec:human-annotation-survey}, we observe that a human would view such an example as a change in vocabulary but with improper usage of the words changed. An example of this type of perturbation is, ``\textit{Tom was a happy man. He lived a simple life.}''. It gets changed to ``\textit{Tom was a grinning man. He lived a bare life.}''

\item \textsc{\textbf{ShuffleSent}}: Important aspects of essay scoring are coherence and organization that measure the extent to which a response demonstrates a unified structure and direction of the narrative. \citep{schultz2013intellimetric,barzilay2008modeling, foltz2013implementation,tay2018skipflow,chen2018end}. To evaluate the dependence of AES scoring on coherence, we randomly shuffle the sentences of a response. This ensures the response's readability and coherence are affected negatively \citep{xu2019cross}. It affects the transition between the lines so that the different ideas appear disconnected to a reader and changes the meaning substantially. 
% \vspace*{-\baselineskip}
\begin{figure}[!ht]
 \centering
 \includegraphics[width=\textwidth]{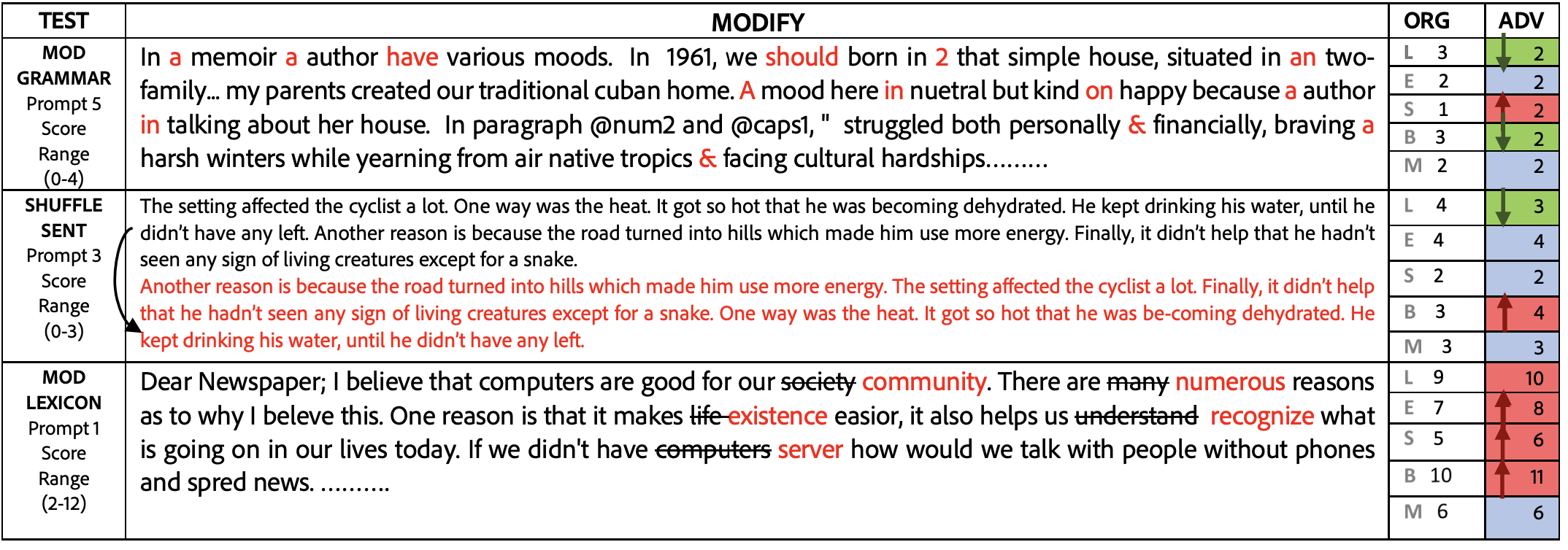}
 \caption{Examples of all the \textbf{\textsc{MOD}} testcases. Here, ORG refers to the score given to the original sample and ADV refers to the score given to the adversarial sample by each model. The list of models according to which scores are listed is LSTM (L), EASE (E), SKIPFLOW(S), BERT (B) and MEMORY NETS (M). \textcolor{red}{Red}, \textcolor{green}{Green}, \textcolor{blue}{Blue} shows that adversarial responses were scored \textcolor{red}{higher}, \textcolor{green}{lower} and \textcolor{blue}{equally} than original response, respectively.}
 \label{modify_examples}
\end{figure}

\end{itemize}

% \vspace*{-\baselineskip}
\subsubsection{\textbf{\textsc{Generative} Adversaries}}
\label{sec:GenerativeAdversaries}

\begin{figure}[!h]
 \centering
 \includegraphics[width=\textwidth]{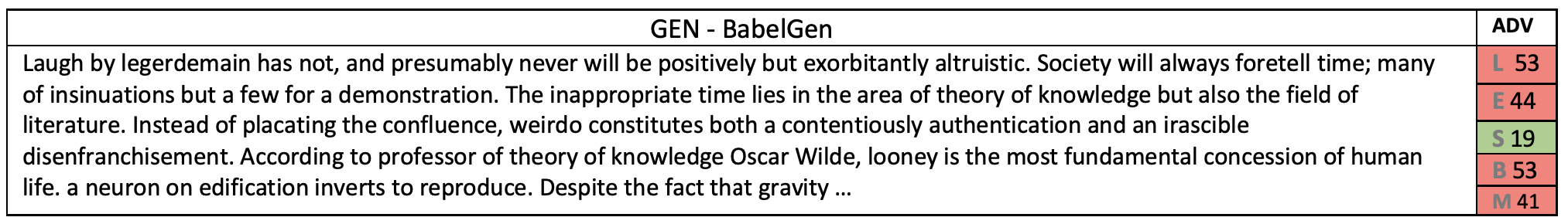}
 \caption{Example of  \textbf{\textsc{GEN}} testcase. Here, ADV refers to the score given to the adversarial sample by each model. BabelGen is used to generate a response using the key words \textit{"laughter", "benefits" and "relationship"} from Prompt 8 (having score range of 0-60). The list of models according to which scores are listed is LSTM (L), EASE (E), SKIPFLOW(S), BERT (B) and MEMORY NETS (M). \textcolor{red}{Red} and \textcolor{green}{Green} show that adversarial responses were scored \textcolor{red}{higher} or  \textcolor{green}{lower} than the mean score (30), respectively.}
 \label{babel_examples}
\end{figure}

%\subsubsection{\textsc{BabelGen}} 
\begin{itemize}
 \item \textsc{\textbf{BabelGen}}: We generate entirely false and gibberish adversarial samples using Les Perelman's B.S. Essay Language Generator (BABEL) \citep{perelmanBable}. BABEL requires a user to enter three keywords based on which it generates an incoherent, meaningless sample containing a concoction of obscure words and keywords pasted together. In 2014, Perelman showed that ETS' \textit{e-rater}, which is used to grade Graduate Record Exam (GRE)\footnote{GRE is a widely popular exam accepted as the standard admission requirement for a majority of graduate schools. It is also used for pre-job screening by a number of companies. Educational Testing Services (ETS) owns and operates the GRE exam.} essays consistently 5-6 on a 1-6 point scale \citep{perelmanBableWebsite,washingtonBabel}. This motivated us to try out the same approach on current state-of-the-art deep learning recent approaches. We came up with a list of keywords based on the AES questions. For generating a response, we chose three keywords as input to BABEL, which then generated a generative adversarial example. Figure~\ref{babel_examples} depicts an example of this testcase.

\end{itemize}

% \begin{figure*}[!ht]

% \centering
% \includegraphics[width=1.0\textwidth]{5-examples.PNG}

% \caption{
% \footnotesize
% Adversarial Samples on different prompts of the types \textsc{AddTruth}, \textsc{RepeatSent}, \textsc{AddSong}, \textsc{AddSpeech}, \textsc{BabelGen}. The (original, adversarial) scores of the different models are: \textsc{AddTruth} (Prompt 5) \{1:(3,2), 2:(2,3), 3:(2,2), 4:(2,1), 5:(3,3)\},
%  \textsc{RepeatSent} (Prompt 7) \{1:(21,19), 2:(18,20), 3:(18,22), 4:(13,14), 5:(17,16)\},
% \textsc{AddSong} (Prompt 6) \{1:(4,4), 2:(2,3), 3:(3,4), 4:(2,2), 5:(2,2)\},
% \textsc{AddSpeech} (Prompt1) \{1:(9,10), 2:(7,8), 3:(5,6), 4:(10,11), 5:(6,6)\},
% \textsc{BabelGen} (Prompt2) \{1:(0,2), 2:(0,4), 3:(0,2), 4:(0,4), 5:(0,5)\}
% }
% \label{fig:some-examples}
% \end{figure*}

%%%%%%%%%%%%%%%%%%%%%%%%%%%%%%%%%%%%%%%%%
%%%%%%%%%%%%%%%%%%%%%%%%%%%%%%%%%%%%%%%%%
%%%%%%%%%%%%%%%%%%%%%%%%%%%%%%%%%%%%%%%%%
%%%%%%%%%%%%%%%%%%%%%%%%%%%%%%%%%%%%%%%%%

\section{Results and Discussion}
\label{sec:Results and Discussion}

% \begin{figure*}[!ht]
% \centering
% \includegraphics[width=0.9\textwidth]{DifferentTestCases.PNG}

% \caption{
% \small{
% Results for \textsc{AddSongs}, \textsc{AddSpeech}, \text{DelRand}, \textsc{AddLies}, \textsc{RepeatSent}, \textsc{AddWikiUnrelated} testcases for two models. Left: Memory Networks \MEMORY, Right: LSTM-MoT \LSTM. The bars above and below x-axis represent $\mu_{pos}$ and $\mu_{neg}$ metrics and the numbers shown on the bars represent $N_{pos}$ and $N_{neg}$ metrics respectively.
% }
% }
% \label{fig:DifferentTestCases}
% \end{figure*}

% ADD A FEW LINES
In this section, we demonstrate our results by performing adversarial perturbations on 2600 original responses and provide a detailed analysis based on our general framework for adversarial evaluation (refer Section~\ref{sec:general-framework}). W
We divide this section into two categories. First, we present the effects on different hyper-parameters such as effect on position, length and amount of change. Secondly, we present results of different test categories (as defined in Table~\ref{table:overall-test-case-list}) such as \textsc{Modify}, \textsc{Add}, \textsc{Delete}, and \textsc{Generate} based Adversaries.

\subsection{Effect of Choice of Different Parameters }
\label{sec:Effect of Choice of Different Hyperparameters}

In this section, we evaluate the effect of various parameters as defined in Section~\ref{sec:general-framework} namely, effect of percentage amount of change $c_1$ of original response by adversary for different values of $c_1$ (\{5, 10, 15, 20, 25\}) and effect of position $c_2$ which defines the position ($\{\textsc{Start, Mid, End}\}$) of inducing adversarial perturbation. 

\subsubsection{\textbf{Effect of Amount of Change ($c_1$)}}
\label{sec:Effect of Amount of Change}
\begin{comment}
STRUCTURE :
1. Only talking about how c1 nad c2 affected the scores.
2. Focus on very slight increase/decrease in scores hence, c1 and c2 not being so significantly imp.
3. Talk about the Prompts specific detail
4. Talk about any peculiar case
5. Talk about best/worst performing model

XXX: Make a table of this format:
Type Subtype Avg change in score Std Dev
 5 10 15 20 25
Add AddTruth x y z w a std_dev(x,y,z,w,a)
\end{comment}

%general trend
% \LSTM 

 %TODO : ABBREVIATION FOR MODELS

\begin{figure}[!h]
    \centering
    \includegraphics[width=\textwidth]{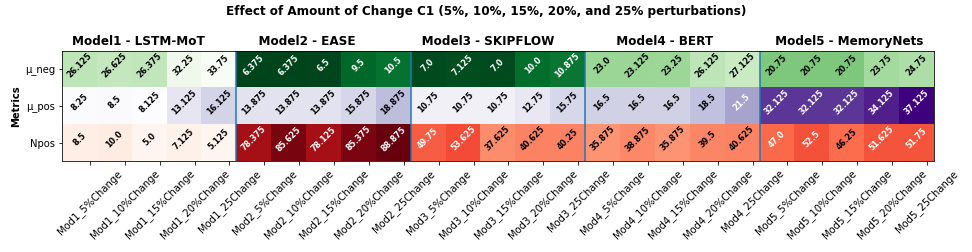}
    \caption{Results for effect of amount of change $c1$ for 5\%, 10\%, 15\%, 20\% and 25\% perturbations. \textsc{Mod1, Mod2, Mod3, Mod4, Mod5} refers to respective Model number.}
    \label{fig:c1Results}
\end{figure}

For various tests listed in Section~\ref{sec:adversarial-evaluation}, we vary the percentage amount of change of perturbation ($c1$) to observe how the model scores such responses. Figure~\ref{fig:c1Results} shows the average difference of scores (averaged over all the different tests). For all models, going from 5\% to 25\% perturbation leads to an increase in of $\mu_{neg}$ and $\mu_{pos}$ of  34\% and 47\%, respectively. We observe that the scoring trend changes considerably while going from 15\% to 20\% perturbation, otherwise it remains consistent.  Exceeding $c1$ to more than 25\% does not add more value to the results. It is clear, irrespective of increase in $c1$, all models except \EASE and \SKIPFLOW have hardly any increase in their $N_{pos}$ value (5\% change). Hence  these models score similar number of responses higher than original but with greater intensity ($\mu_{pos}$) as the amount of perturbation increases. We infer that these models are overstable with respect to number of adversarial responses they are scoring higher or lower, due to the consistent value of $N_{pos}$ with an increase in $c1$. It is unexpected to see that \EASE has scored an average of 82\% adversarial responses higher than the original ($N_{pos}$). This value is the lowest for \LSTM, averaging to only 7\%.
% \vspace*{-\baselineskip}

\subsubsection{\textbf{Effect of Position Criterion ($c_2$)}} 
\label{sec:Effect of the bounded size criterion $c_2$}

\begin{comment}
1. which prompt has max deviation for all models XXX, XXX, XXX, XXX, XXX (mention both upos and uneg values). Hence the overall average is XXX for upos and XXX for uneg. 
2. why does this happen(add subjective line)
3. overall max deviation overall all prompts for each prompt is XXX\% for upos/uneg. mention about standard deviation too. Highest is for dash model is XXX\% which means that XXX\% of the samples 
\end{comment}

We perform an analysis to show whether the addition of content at specific positions namely ({\textsc{Start}, \textsc{Mid}, \textsc{End}}) under the following conditions.
\begin{itemize}
    \item \textit{bounded:} retaining the length of the response after the addition of content.
    \item \textit{unbounded:} no restrictions on the length of the response despite the addition of content.
\end{itemize}

% In Figure~\ref{position_allModels}, we observe that the scores are not affected concerning the position. However, for model \LSTM, it can be observed that the addition of lines, in the beginning, has increased the $N_{pos}$ metric by an average of 7\%. This implies that more adversarial responses are scored higher when lines are added in the starting of the response compared to middle or end, for this model. 

\begin{comment}
THIS DOES NOT GIVE ANY SIGNIFICANCE TO BOUNDED/UNBOUNDED.... maybe do a separate section for this?>

Firstly, we observe that argumentative and narrative-based prompts (Prompts 1, 2, and 7, 8 respectively) show the least deviation in scores. For model \EASE, we see a low $\mu_{pos}$ of 4\% in average and low $\mu_{neg}$ of 3\% for these prompts. Additionally, model \SKIPFLOW indicates a similar fashion with $\mu_{pos}$ around 6\% and $\mu_{neg}$ around 4.7\%. As the intensity of change in scores is small (less than 5\%) for both models, we conclude that they maintain the scores of unmodified original responses even after adversarial modifications. This indicates the \textit{overstability} of models \EASE and \SKIPFLOW for position criterion $c_2$, as they are not able to distinguish ill-formed responses from the well-formed ones.
\end{comment}

\begin{figure}[!ht]
    \centering
    \includegraphics[width=\textwidth]{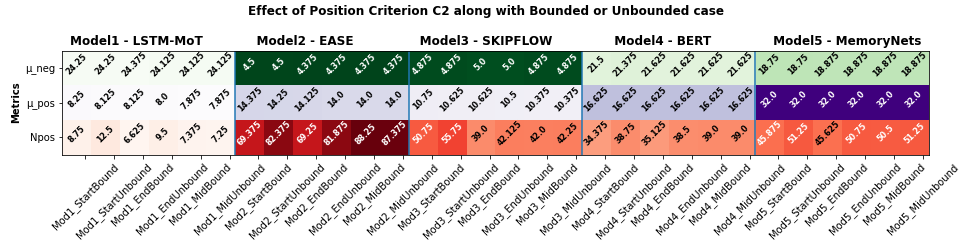}
    \caption{Results for Effect of position criterion $c_2$ for bounded and unbounded case across three positions \textsc{Start}, \textsc{End} and \textsc{Mid}. \textsc{Mod1, Mod2, Mod3, Mod4, Mod5} refers to respective Model number.}
    \label{fig:c2Results}
\end{figure}

% \textbf{Variation with respect to the length and the position of perturbation: } 
Across all models as demonstrated in Figure~\ref{fig:c2Results}, we see an equivalent variation of $\mu_{pos}$ and $\mu_{neg}$ values for both bounded and unbounded cases, irrespective of the position and length of the perturbation. 
% This implies that the intensity with which adversarial responses are scored positively or negatively do not differ for bounded or unbounded cases, irrespective of the position in which the perturbation is added. 
However, we see that $N_{pos}$ increases for unbounded cases when compared to bounded cases for all test cases across all models. For the \textsc{START} position, we see an increase of 12\% on an average in $N_{pos}$ from bounded to the unbounded situation. This means that models are sensitive towards an increase in the number of words by scoring more number of responses higher but with similar intensities. We observe a similar trend for \textsc{END} position but with a lower $N_{pos}$ increase of 7\% in average. We can say that scores are proportional to the length of the response for the \textsc{START} and \textsc{END} positions. However, addition in \textsc{MID} position does not influence the scores differently based on the length of the essay. This resonates with \citep{perelman2014state} as he states that word count is the most important predictor of an essay's score. From Figure~\ref{fig:c2Results}, we also notice the intensity of change in scores is small (less than 5\%) for models \EASE and \SKIPFLOW. Both are \textit{overstable} concerning position criterion $c_2$, as they are not able to distinguish ill-formed responses from the well-formed ones. Even when all prompts are taken into consideration, the deviation for these models are 8\% and 9\% respectively, reinforcing the previous insight. However, the model \EASE has the highest number of positively affected adversaries (80.3\% of responses). This is unexpected as perturbation should decrease the scores of the responses. Amongst the other models, we observe that model \MEMORY is the worst performing model as it detects 49.2\% (approximately one half) with a rise in score by 23.5\% and the other half of responses with a fall with the same intensity. This shows the model does not know which direction to move when scores are changed. On the contrary, the model \LSTM is the best performing model considering all adversarial evaluation metrics. 
\subsection{Results of the different types of Adversaries}
\label{sec : Results of the different types of Adversaries}

\subsubsection{\textbf{\textsc{Add} Adversaries }}
\label{sec:Addition_Adversaries}

\begin{figure}[!ht]
\begin{minipage}{\textwidth}
\centering
\subfloat{\label{main:a}\includegraphics[scale=0.42]{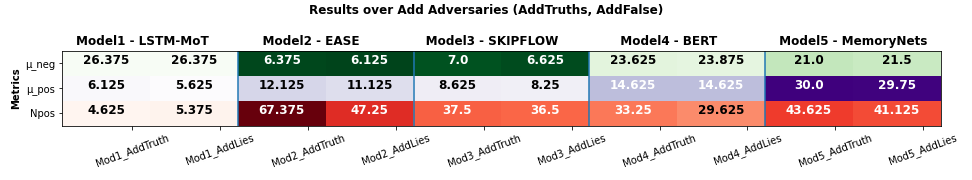}}
\end{minipage}%
% \begin{minipage}{\textwidth}
% \centering
% \subfloat{\label{main:b}\includegraphics[scale=1]{example-image-b}}
% \end{minipage}
\par\medskip
\centering
\subfloat{\label{main:b}\includegraphics[scale=0.42]{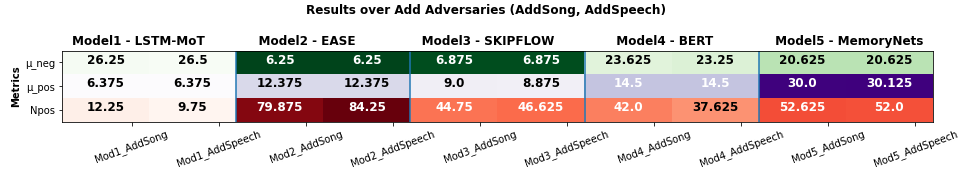}}
\par\medskip
\centering
\subfloat{\label{main:c}\includegraphics[scale=0.42]{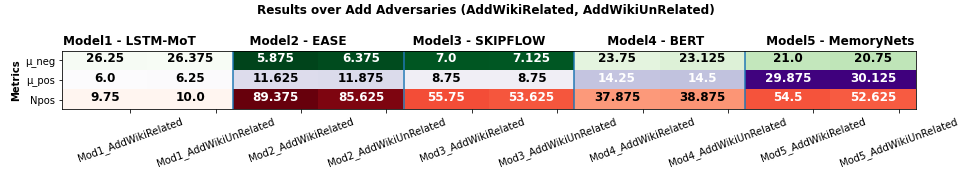}}

\caption{Results over Add Adversaries. \textsc{Mod1, Mod2, Mod3, Mod4, Mod5} refers to respective Model number.}
\label{fig:AddResults}
\end{figure}

In this section, we explain the results over all the tests for \textsc{Add} adversaries (refer Section~\ref{sec:Add Adversaries}). We show all adversarial evaluation metrics (refer Table~\ref{table:adversarial-eval-metrics} and Figure~\ref{fig:AddResults}) for all the models (refer Section~\ref{subsection:Models}). 
We observe that that two out of five models, \EASE and \SKIPFLOW, show the least $\mu_{pos}$ (12\% and 8.7\% respectively) and $\mu_{neg}$ (6.2\% and 6.9\% respectively) after adversarial modifications. We infer that these models are overstable as the intensity of change in scores is small for both the models, and they score the adversarial changed responses similar to that of unmodified original responses. Model \MEMORY unexpectedly scores about 50\% of the adversarial responses higher than original with high $\mu_{pos}$ of about 30\%. In contrast, \LSTM has scored the adversarial responses in a highly negative fashion. The value of $N_{pos}$ is only 8.6\%. This implies that 91.4\% of modified responses are scored lower than their respective original response. Moreover, the value of $\mu_{neg}$ is 27\% while $\mu_{pos}$ is only 6.1\% over all the prompts. This symbolizes that this model can observe perturbations in the responses and score them relatively lower. Hence, amongst all the models, these two show relatively better performance. It is interesting to observe that out of all testcases, the test \textsc{AddLies} has around 50\% $N_{pos}$ for the models \EASE and \MEMORY. These models are not able to penalize the deliberately included false facts into the response. However, we observe that \textsc{AddTruth} (as shown in Figure \ref{fig:AddResults}) is scored comparatively higher by all models. On a relative note, false statements have impacted scores negatively, even if they do so marginally. We believe this is because most models used contextual word embeddings as inputs to their models.  Mostly, we notice the tendency of lengthier responses to be scored higher, despite being factually wrong and having unrelated content. Ideally, we felt that added irrelevant lines from songs, speeches, and Wikipedia articles would likely make the models score the responses lower than the addition of relevant content. However, these test cases were scored no differently than the rest, suggesting that addition of relevant or irrelevant lines were both scored in similar manner. This means that the models do not check for the relevance and sufficiency content features of an essay, which should play an important role in scoring \cite{Gusfield:97}.
% \vspace*{-\baselineskip}

\subsubsection{\textbf{\textsc{Delete} Adversaries}}
\label{sec:Delete Adversaries}
% In general, \textsc{DELETE} tests impacted the scores negatively. There were very few instances where scores increase after removing a few lines. We can spot this over all prompts and across all models. 

\begin{figure}[!ht]
    \centering
    \includegraphics[width=\textwidth]{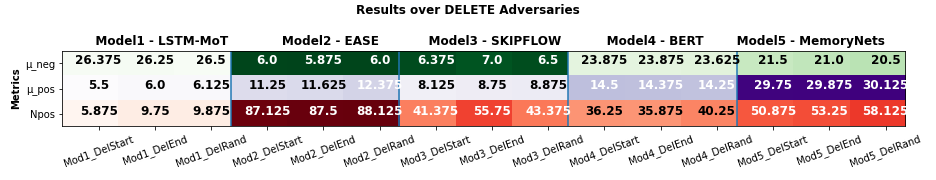}
    \caption{Results over \textsc{Delete} Adversaries. \textsc{Mod1, Mod2, Mod3, Mod4, Mod5} refers to respective Model number.}
    \label{fig:DelResults}
\end{figure}

This section describes the results over all tests for \textsc{Delete} Adversaries (refer Section~\ref{sec:Del Adversaries}). We demonstrate all adversarial evaluation metrics (refer Table~\ref{table:adversarial-eval-metrics} and Figure~\ref{fig:DelResults}) over all the models (refer Section~\ref{subsection:Models}). 

We find that model \EASE and \SKIPFLOW show the least $\mu_{pos}$ and $\mu_{neg}$ values after adversarial modifications. This means that both models are hardly fluctuating from the original scores of unmodified responses. This indicates characteristics of \textit{overstability} of both models. In model \MEMORY, we see high values of $\mu_{pos}$ and $\mu_{neg}$, aggregating with a high $N_{pos}$ value of 54\%, in average. In other words, the model is scoring half of the responses higher, with an average of 30\% soar ($\mu_{pos}$) and scoring the other half lower, with a dip of 22\% ($\mu_{neg}$). This means that the model is responsive to \textsc{Delete} adversaries but in no particular direction. Model \LSTM has scored adversarial responses majorly in a negative fashion by observing the highest average $\mu_{neg}$ value of 26.4\% when compared to a low $\mu_{pos}$ of 5.9\%. Additionally, We calculate the $N_{pos}$ to be 8.5\%, which means that 91.5\% of samples have been scored negatively. We draw the inference that this model can observe the presence of adversarial perturbations in the responses. Moreover, we mark a similar trend for model \BERT, however, with higher intensities of average deviation ($\mu_{pos}$ and $\mu_{neg}$) in the scores. We summarize that model \LSTM is the best performing model. Looking into testcase based results, we see that \textsc{DelRand} has high $N_{pos}$ and $\mu_{pos}$ (increase of 2\% and 3\% )for as compared to \textsc{DelStart} and \textsc{DelEnd} tests. This implies that adversarial responses where the lines were randomly deleted were positively scored with a higher $\mu_{pos}$ than those responses in which the introduction and conclusion was removed. This is surprising as deletion of random lines from the response leads to a loss in organization and response structure. Deletion at the end has a higher $N_{pos}$ (more that 50\% of the responses scored higher than original) on average for three out of five models. The responses in this test case were missing any concluding remarks. Hence, the capability of models to check for a proper conclusion is poor. 

\subsubsection{\textbf{\textsc{Modify} Adversaries}}
\label{sec:Modify_Adversaries}

\begin{figure}[!ht]
    \centering
    \includegraphics[width=\textwidth]{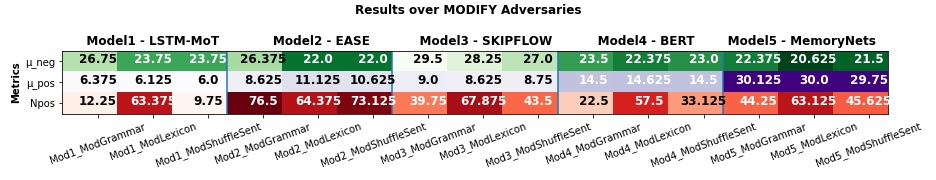}
    \caption{Results over \textsc{Modify} Adversaries. \textsc{Mod1, Mod2, Mod3, Mod4, Mod5} refers to respective Model number.}
    \label{fig:ModifyResults}
\end{figure}

This section, explains the results over all tests for \textsc{Modify} based Adversaries (refer Section~\ref{sec:Modify Adversaries} ). We depict the adversarial evaluation metrics (mentioned in Table~\ref{table:adversarial-eval-metrics} and Figure~\ref{fig:ModifyResults}) for all models (refer Section~\ref{subsection:Models}).

% From Figure~\ref{fig:ModifyAll}, we find among all the models, \EASE shows the least $\mu_{pos}$ (10\%) and $\mu_{neg}$ (20\%) after adversarial modifications. Moreover, $\mu_{pos}$ is less than 4\% on average for argumentative and narrative-based prompts (Prompts 1, 2 and 7, 8 respectively). We infer that the model is \textit{overstable}.

We observe that Models \MEMORY and \EASE and \SKIPFLOW has $N_{pos}$ greater than 50\% of the total number of responses. That means more than half of the responses have been scored higher than original responses. \textsc{Modify} test cases such as \textsc{ModGrammar} and \textsc{ShuffleSent} significantly affects the discourse of the response in a negative manner and also makes it unorganized and unstructured. Hence, ideally these responses should not have been scored positively. This shows that these models are not able to capture the discourse and organization based relevance of the responses.On the other hand, we observe that the model \LSTM has scored the adversarial responses in a highly negative fashion. These responses generally scored lower (89\% modified responses are scored lower than their respective original response) and with high intensity, as shown by $\mu_{neg}$ of 23\%. Over all eight prompts, we see that greater values of $\mu_{neg}$ compared to only 6\% value of $\mu_{pos}$. This symbolizes that this model has the ability to act robustly in the presence of adversaries. Among all \textsc{Modify} Adversaries (refer Section~\ref{sec:Modify Adversaries}), we observe that test \textsc{ModGrammar} had a consistently low score amongst all the models. This can be verified as the measure $N_{pos}$ is significantly lower in all the models except \EASE. Overall, $N_{pos}$ constitutes of only 36\% of all the adversarial responses. This shows that most models can identify grammatically incorrect sentences and score them lower. The intensity of scoring grammatically incorrect adversarial responses negatively is also higher than that of\textsc{ModLexicon} and \textsc{ModShuffle}. However, for model \EASE the trend is opposite with respect to $N_{pos}$. An average of 83\% of incorrect grammar adversarial responses are scored positively, in this case. This shows that model \EASE has problems recognizing grammatical errors in the responses. Moreover, it is scoring these adversarial responses higher than the original. Again, \LSTM has correctly scored most of the \textsc{ModGrammar} and  \textsc{ModShuffle} lower than \textsc{ModLexicon} (Figure \ref{fig:ModifyResults}), which is how we expect all models to infer these testcases.

\subsubsection{\textbf{\textsc{Generate} Adversaries}}
\label{Sec:Generate_Adversaries}

Another category of test case \textsc{BabelGen} where we generate incoherent and meaningless responses. Ideally, this should have been scored a zero but as demonstrated in Table~\ref{table:Babel-results}, we notice that almost all the models score these generated essays at least 60\% of the prompt scoring range. This strongly suggests that models were looking for obscure keywords with complex sentence formation. We can also infer that the relevance of the responses with respect to the question is missing. Since the responses are generated using key words, they contain sentences with respect to those key words, but fail to answer the question targeted. 

\begin{table}[htbp]
\footnotesize
\centering
\begin{tabular}{lllllllll}
\hline
\textbf{M/P} & \textbf{1} & \textbf{2} & \textbf{3} &\textbf{ 4} & \textbf{5} &\textbf{ 6 } &\textbf{ 7 } & \textbf{8 } \\\hline
\small

Range & 2-12 & 1-6 & 0-3 & 0-3 & 0-4 & 0-4 & 0-30 & 0-60 \\ \hline
1 & 7.1 & 2.5 & 1.7 & 1.1 & 2.2 & 1.2 & 13.8 & 33.9 \\
2 & 10 & 4.4 & 2 & 2 & 3 & 1.2 & 19.1 & 43.1 \\
3 & 6 & 2 & 1.1 & 0.9 & 1.3 & 1.3 & 12.1 & 21.9 \\ 
4 & 8.4 & 4 & 3 & 3 & 4 & 3.9 & 18.4 & 40.1 \\ 
5 & 10.8 & 5.6 & 2.8 & 2.9 & 3.8 & 3.8 & 26.2 & 53 \\ \hline
\end{tabular}
\caption{\label{table:Babel-results} 
\small
Scores for \textsc{BabelGen} over all the prompts and models. Ideally, all of the Babel generated essays should have been scored a zero.
Legend :  M: Model (y-axis), P: Prompt (x-axis), Model Types:
1:\LSTM, 2:\EASE, 3:\SKIPFLOW, 4:\MEMORY, 5:\BERT.
}
\end{table}

% Newpage
% \begin{figure*}[!ht]
% \centering
% \includegraphics[width=0.9\textwidth,scale=0.8]{adversarial_training.PNG}

% \caption{
% \small{
% Results of adversarial training for Prompts 2,3,5,7 in clockwise order. The x-axis shows chosen test-cases and y-axis shows 4 metrics: \{$\mu_{pos}, \mu_{neg}, N_{pos}, \sigma$\}. Representations: The solid lines denoted by $metric_{same}$ : Value of $metric$ with adversarial training done over the data generated by the same test case, the dashed lines denoted by $metric_{diff}$ : Value of $metric$ with adversarial training done over the data generated by a different test case, the dotted lines denoted by $metric_{original}$ : Value of $metric$ with no adversarial training done
% }
% }
% \label{fig:adv-training}
% \end{figure*}
% Another category of test case \textsc{BabelGen}. Ideally, this should have been scored a zero but almost all the models scored at least 60\% to the generated essays. This strongly suggests that models were looking for obscure keywords with complex sentence formation. We also observed that modifying grammar did not affect the scores much or affected it negatively. This is largely in congruence with the rubrics of the questions where it was indicated that grammar should not be valued for scoring. However, unexpectedly, in some cases after changing the grammar of the whole response, we observed that scores started increasing.
%A few examples demonstrating this are given in the supplementary.
% \vspace*{-\baselineskip}
% \vspace*{-1mm}
\subsection{Human Annotation Survey}
\label{sec:human-annotation-survey}
% \vspace{-3 pt}
% \vspace*{-\baselineskip}

\begin{figure}[!ht]
\centering
\includegraphics[width=0.8\textwidth]{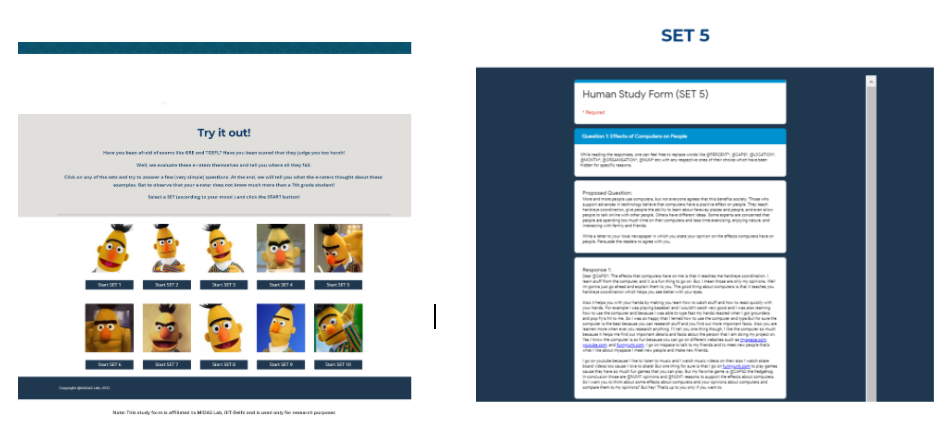}

\caption{
\small{
Glimpses from our human survey website.
}
}
\label{fig:screenshot-survey}
\end{figure}

We conducted a social survey with 200 participants to understand and compare how humans score our tests compared to the automatic essay scoring systems. Figure~\ref{fig:screenshot-survey} show a few screenshots from our survey website. To create our survey forms, we chose test cases based on the following three conditions: 1) where $N_{} < N_{neg}$, 2) where $\mu_{pos} > 10\%$, 3) where a T-test rejects the hypothesis that the adversarial and original scores are the same distribution and 4) . The motivation behind setting these three conditions was that we wanted to choose those test-cases where the model should be most confident in scoring adversarial response as negative and unfavorable. Once annotated by humans, we show that these systems, we compare the differences in these scores. We observe the AES systems lack the ability to \textit{adequately} penalize scores by either marking marking the perturbations as better than the original ($N_{} > N_{neg}$) or not detecting any significant difference. Both are wrong presumptions by the model.

\begin{figure}[!ht]
\centering
\includegraphics[width=0.7\textwidth]{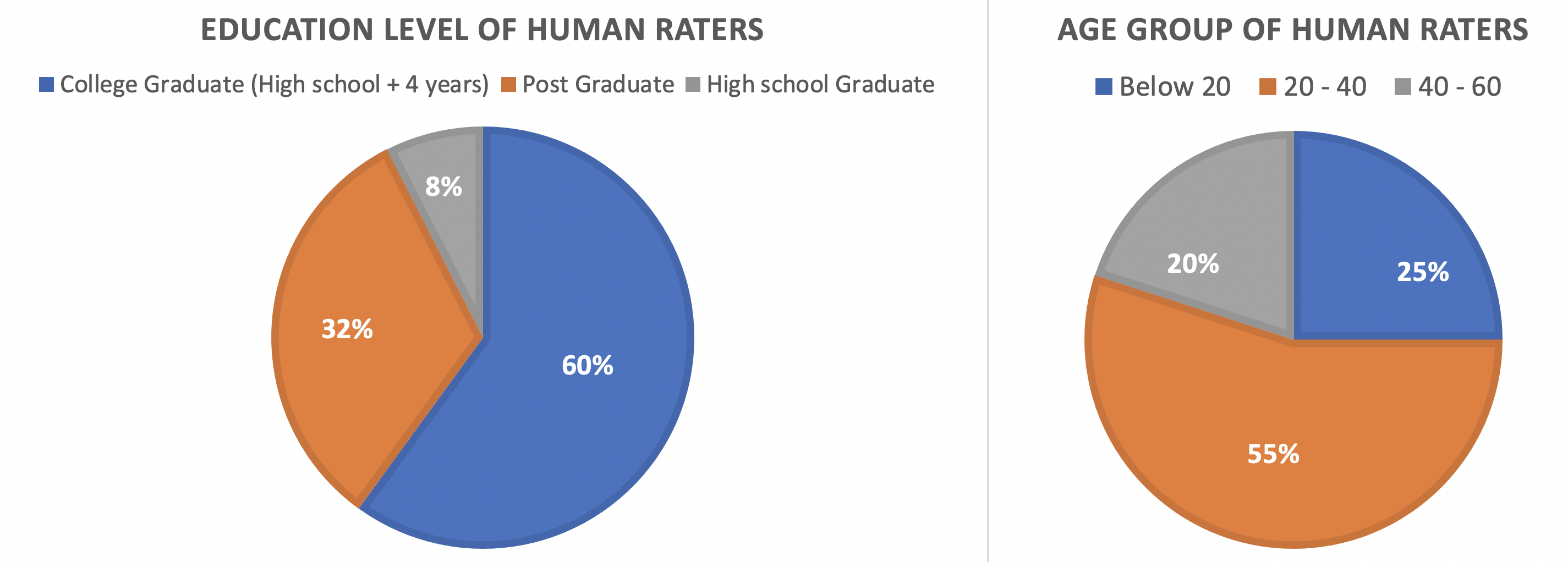}

\caption{
\small{
Estimate of the education level and age group of 200 human raters who participated in the survey.
}
}
\label{fig:pie-chart}
\end{figure}
%If the human annotations showed that even on these test-cases where model is confident of the perturbation being negative and yet not able to adequately penalize the adversarial samples, the other test-cases are automatically validated since models are either marking the perturbations as better than the original response or not detecting any significant difference. 
Table~\ref{table:HumanAnnotations} depicts the results of our human annotations. We divide the annotators into two groups. We show them the original response and its corresponding score for the first group and then ask the annotators to score the adversarial response accordingly. For the second group, we ask them to score both the original and adversarial responses. If any of the annotators felt that both the responses' scores should not be the same, we ask them to list supporting reasons. For uniformity in responses, we derive a set of scoring rubrics, mentioned in our dataset and ask them to choose the most suitable keywords. As observed from Table~\ref{table:HumanAnnotations}, the percentage of people who scored adversarial responses \textbf{lower} than original responses is significantly higher for all selected test-cases. The main reasons for scoring adversarial responses lower by annotators are \textit{Relevance, Organization, Readability, etc}. It can be observed that the percentage lowering in score was on an average of 30\%. 

\begin{table*}[!ht]
\resizebox{\textwidth}{!}{
\begin{tabular}{lllllll}
\hline
\textbf{\#} & \textbf{Perturbation} & \textbf{Score $\downarrow$ \%} & \textbf{\% People $\downarrow$} & \textbf{\% People $\uparrow$} & \textbf{Common Reasons of $\downarrow$} & \textbf{Common Reasons of $\uparrow$} \\\hline

%1 & \textsc{ModFluency} & 28.1 & 82.7 & 4.8 & { \small Std English, Readability} & {\small More appropriate} \\
1 & \textsc{Shuffle} & 24.2 & 68.6 & 14.5 & Transitions ,Organization, Relevance & None\\
2 & \textsc{ModGrammar} & 39.5 & 91.3 & 6.2 & {Grammar, Conventions, Readability} & None\\
3 & \textsc{AddWikiRelated} & 38.2 & 87.2 & 11.3 &Readability, Relevance, Conventions & Transitions\\
4 & \textsc{RepeatSent} & 15.6 & 71.6 & 13.6 &Organization, Relevance, Repetition &Clarity\\
5 & \textsc{AddLies} & 23.9 & 79.9 & 10.6 &Relevance, Organization &Conventions\\
6 & \textsc{AddTruth} & 29.2 & 88.6 & 8.6 &Relevance, Readability &Organization\\
7 & \textsc{AddSong} & 32.8 & 91.8 & 3.2 & {Relevance, Organization, Grammar} & {Both equal} \\
8 & \textsc{DelRand} & 38.2 & 87.2 & 11.3 & {Transitions, Organization} & {Same, More appropriate} \\ \hline

\end{tabular}
}
\caption{Human Annotation Survey Results. ($\downarrow$ represents a decrease and $\uparrow$ represents an increase. Therefore, `\%~People~$\downarrow$' denotes the percentage of people who scored the adversarial response worse than the original response and `Score$\downarrow\%$ ' corresponds to the average percentage amount the new score dropped from the original score of the response.)}
\label{table:HumanAnnotations}

\end{table*}
% \begin{table*}[]
% \resizebox{\textwidth}{!}{\begin{tabular}{l|lll|lll|lll|lll}
% \hline
% \textbf{Test-case} & Prompt 2 & & & Prompt 3 & & & Prompt 5 & & & Prompt 7 & & \\\hline
% \textbf{Metric -\textgreater{}} & u\_neg & u\_ & std & u\_neg & u\_ & std & u\_neg & u\_ & std & u\_neg & u\_ & std \\ \hline
% incorrect\_grammar & 0 & 1.706 & 0.094 & -0.660 & 0.865 & 0.639 & -0.411 & 1.156 & 0.558 & -3.429 & 7.045 & 4.768 \\
% shuffle & -0.711 & 1.166 & 0.273 & -0.499 & 1.138 & 0.50 & -0.770 & 1.054 & 0.891 & -3.360 & 9.800 & 3.927 \\
% speech\_mid & -0.458 & 1.654 & 0.540 & -0.682 & 1.086 & 0.566 & -0.748 & 1.360 & 0.635 & -2.434 & 10.075 & 4.222 \\
% uf\_mid & -0.738 & 1.477 & 0.963 & -0.280 & 1.294 & 0.384 & -0.847 & 1.527 & 0.77 & -3.622 & 8.601 & 5.704 \\
% wiki\_beg & -0.253 & 2.237 & 0.300 & -0.507 & 1.180 & 0.473 & -0.611 & 1.265 & 0.537 & -3.97 & 6.830 & 4.897\\ \hline
% \end{tabular}}

% \label{adversarial-training}
% \caption{Adversarial Training Results}
% \end{table*}
\begin{figure}[!ht]
\centering
\includegraphics[width=\textwidth]{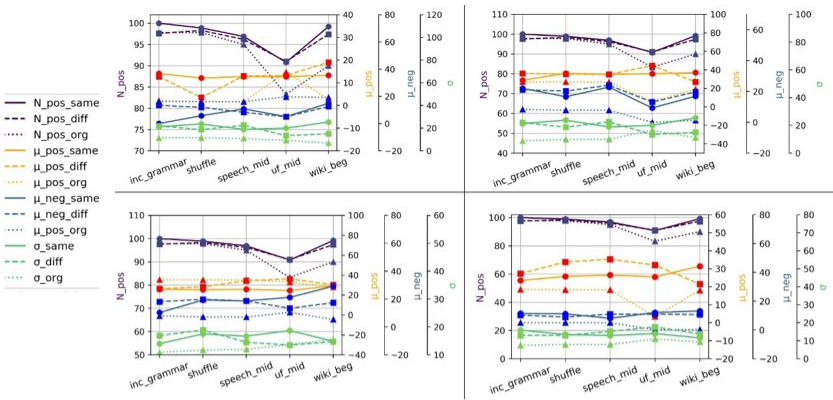}

\caption{
\small{
Results of adversarial training for Prompts 2,3,5,7 in clockwise order. The x-axis shows chosen test-cases and y-axis shows 4 metrics: \{$\mu_{pos}, \mu_{neg}, N_{pos}, \sigma$\}. Representations: he solid lines denoted by $metric_{same}$ : Value of $metric$ with adversarial training done over the data generated by the same test case, the dashed lines denoted by $metric_{diff}$ : Value of $metric$ with adversarial training done over the data generated by a different test case, the dotted lines denoted by $metric_{original}$ : Value of $metric$ with no adversarial training done.
}
}
\label{fig:adv-training}
\end{figure}
\vspace*{-\baselineskip}
\subsection{Adversarial Training}
\label{sec:Adversarial Training}
%\vspace{-3 pt}

Finally, we performed an experiment by training on the adversarial samples generated by our framework to see if the models can pick up some inherent ``pattern'' of the adversarial samples. Since there is a multitude of adversarial test cases category, we narrowed a subcategory of five test cases from those shown for the human annotations. They were selected such that on an average, these test cases had maximum deviation between human annotated scores and machine scores. The train data consisted of an equal number of original samples and adversarial samples. The target scores of adversarial samples were set as the original score minus the mean difference of scores between original and human-annotated scores. For example, according to the human annotation study, for the \textsc{ModGrammar} case, the mean difference was 2 points below the original score, so all the samples were scored as original scores minus 2 points in the simulated training data. The simulated training data was then appended with original and shuffled. The testing was conducted with the respective adversarial test-case as well as the others. The results for the same is shown in Figure~\ref{fig:adv-training}. It is evident that the adversarial training improves the scores marginally for all four metrics, as shown by the solid lines being higher than the dotted lines. However, a slightly visible improvement in scores is inapparent. The $N_{pos}$ increases for adversarial training, highest for the respective test-case. Similar trend is observed for $\mu_{neg}$ metric. For $\mu_{pos}$, the adversarial training reduces this score for respective testcase, as compared to non-adversarial testing.

% \vspace*{-\baselineskip}
\section{Conclusion and future work}
\label{sec:Conclusion}
%\vspace{-4 pt}
Through our experiments. we conclude that current AES systems built mainly with feature extraction techniques and deep neural networks based algorithms fail to recognize the presence of common-sense adversaries in student essays and responses. As these common adversaries are popular among students for `\textit{bluffing}' during examinations, it is vital for Automated Scoring system developers to think beyond the accuracies of their systems and pay attention to complete robustness so that these systems are not vulnerable to any form of adversarial attack. The future scope of this work includes designing more efficient AES systems using the metrics proposed, combining the metrics to provide a more holistic criteria for analysis and improving the evaluation suite with emphasis to the type of exam and level of education of the students.

 \bibliographystyle{ACM-Reference-Format} % basic style, author-year citations

\bibliography{bibFile.bib} % name your BibTeX data base

\end{document}